# Mechanisms of Artistic Creativity in Deep Learning Neural Networks

**Lonce Wyse**
Communication and New Media Department
National University of Singapore
Singapore
lonce.wyse@nus.edu.sg

**Abstract**

The generative capabilities of deep learning neural networks (DNNs) have been attracting increasing attention for both the remarkable artifacts they produce, but also because of the vast conceptual difference between how they are programmed and what they do. DNNs are "black boxes" where high-level behavior is not explicitly programed, but emerges from the complex interactions of thousands or millions of simple computational elements. Their behavior is often described in anthropomorphic terms that can be misleading, seem magical, or stoke fears of an imminent singularity in which machines become "more" than human.

In this paper, we examine 5 distinct behavioral characteristics associated with creativity, and provide an example of a mechanisms from generative deep learning architectures that give rise to each these characteristics. All 5 emerge from machinery built for purposes other than the creative characteristics they exhibit, mostly classification. These mechanisms of creative generative capabilities thus demonstrate a deep kinship to computational perceptual processes. By understanding how these different behaviors arise, we hope to on one hand take the magic out of anthropomorphic descriptions, but on the other, to build a deeper appreciation of machinic forms of creativity on their own terms that will allow us to nurture their further development.

## Introduction

Deep Learning neural networks are notorious for their opacity. We know exactly what is happening at the level of activations and weighted connections between the millions of nodes that may comprise a network, but we are lacking in the analytical tools that would provide human-understandable explanations for decisions or behavior. "Explainable AI" is a prominent goalpost in current machine learning research, and in particular, within the field of Computational Creativity ("CC") (Bodily and Ventura 2018). Guckelsberger, Salge, and Colton (2017) and others have proposed that the system itself should have a reflective understanding of why it makes creative decisions in order to be considered creative. By understanding the mechanisms that give rise to creative behavior, we might better be able to build systems that can reflect and communicate about their behavior.

"Mechanisms" have a contested status in the field of computational creativity. Fore some, there is a reluctance to make the recognition of creativity depend on any particular mechanism, and focus is directed to artifacts over process (Ritchie 2007). This at least avoids the phenomenon plaguing classically programmed AI systems identified by John McCarthy that, "as soon as it works, no one calls it AI any more." Others (Colton 2008) argue that understanding the process by which an artifact in generated is important for an assessment of creativity.

With neural networks, complex behaviors emerge out of the simple interaction between potentially millions of units. The architectures and unit-level learning algorithms in combination with exposure to data for training allow the networks to configure themselves to exhibit behaviors ranging from classification to media generation. These configurations and behaviors are the "mechanisms" that will be used to explain capabilities that appear akin to human creativity. When mechanisms are emergent rather than explicitly programmed, then their "hidden" nature is not just a methodological choice for framing a definition or establishing criteria for creativity. Understanding how they work requires scientific investigation and is less likely to change our interpretation of their behavior than is unveiling an explicit piece of computer code that generates creative behavior.

The emergent nature itself of neural network mechanisms is relevant to current trains of thought in the CC community. Bodily and Ventura (2018) assert that an autonomous aesthetic, not programmed in by a designer, is fundamental to creative systems. Neural networks also foreground the relationship between perception and generation that has deep roots in psychology (Flowers and Garbin 1989), but has sometimes been neglected in classical AI approaches to programming generative systems. This is because many generative neural networks create media based on the very same mechanisms configured during training on input data from the same domain.

Mechanism is also interesting in complex systems because of the way it can play a role in different kinds of behavior. The mechanism that allows a system to see edges across luminance contrast may also cause illusions in re-sponse to particular stimuli. Similarly, learning mechanisms that support generalization from scant evidence might enable taking reasonable novel action under unseen conditions in one context, but be recognized as biased decision making when deployed in an inappropriate context.

In this paper, mechanisms in deep learning architectures that give rise to the following 5 characteristics associated with creativity are explored:

- Transformative perception
- Synthesis of different domains
- Sentiment recognition and synthesis
- Analogic and metaphorical reasoning
- Abstraction

These characteristics are a subset of "sub-models of notions used to describe creativity" in Pease and Colton (2011). For the purposes of this paper, no definition of creativity is attempted or necessary, nor are any claims made about whether the mechanisms discussed are "authentically" creative in a human sense. Anthropomorphic terms will be given mechanistic explanations.

## Transformative perception

"Make it new" is a phrase Ezra Pound canonized, which refers to a transformative way of seeing that lies at the heart of modern artistic processes. Can neural networks see the world in a unique way based on their own individual experience and share their vision with others?

Deep neural architectures are typically layered, with groups of neurons in one layer connected to the next through "synaptic" weights. A typical example is an image classifier where network inputs are pixel values of images, and output layer cell activations are interpreted as class labels.

In an attempt to understand and characterize what the units and layers are actually computing in terms of the input, a procedure known as activation maximization has been developed (Erhan et al. 2009; Nguyen et al. 2016). The method is very similar to those used to probe human neural response characteristics. Hidden units are monitored, and the input images are manipulated in order to amplify the response patterns of particular individuals or ensembles of units. Systematic exploitation of this idea yielded one of the most prominent image-generating deep learning systems to attract widespread attention beyond the research community, the DeepDream network (Mordvintsev et al. 2015). The manipulated images produced in the service of understanding response characteristics of network units turned out to be striking for their hallucinogenic and dream-like character, and thus an artistic generative technique was born.

Architectures such as the DeepDream network also afford a systematic way to "parametrically" control certain characteristics of the images generated in this way. It turns out (again in a loose analogy to the way the human brain is structured), that peripheral layers tend to respond to "low level" features such as edges, while units in deeper layers tend to respond to higher-order features such as spatial relationships between lower level feature patterns. As deeper layers are probed, the nodes respond to what we recognize as semantic features. When images are manipulated to maximize the activation level of a particular hidden unit (or ensemble), the results are dependent on the nodes and layers being probed. Images manipulated to maximize activity in peripheral layer nodes yield edge (and other low-level feature) enhancements (Figure 1), while images manipulated to maximize the response of units in deeper layers reveal patterns related to the objects that the networks were trained to categorize at their output.

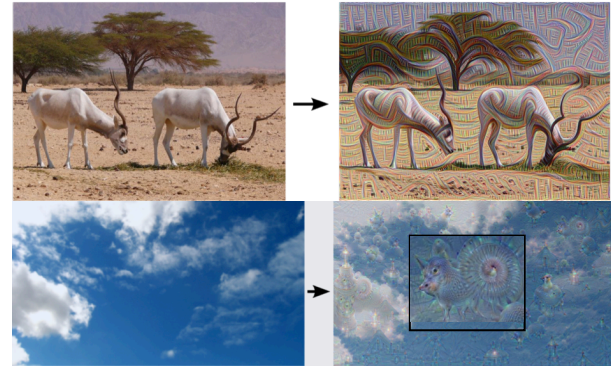

Figure 1. Top: One of the original images presented to the DeepDream network, and the image systematically transformed to enhance highly activated nodes in a peripheral layer. Bottom: An original and a transformed image that enhance highly activated nodes in a deep layer. Inset: A zoomed section of transformed image. (Mordvintsev et al. 2015)

This example illustrates the close relationship between the perceptual and generative capabilities of this kind of deep learning architecture. This is because the novelty in the generated image arises from the technique of exploiting the knowledge in weighted connections learned during perceptual training on input images in order to "read into" the peripheral image during generation. This machinic process of apophenia has been interpreted in other contexts as a dangerous tendency of neural nets to be biased (Steyerl 2018) by interpreting their environments based on their previous limited exposure to data rather than objectively. Apophenia can also be seen as a kind of cross-domain synthesis between a specific image and those previously seen and learned, but a more specific form of cross-domain synthesis will be discussed in the next section.

## Cross-domain synthesis

We commonly recognize a distinction between content and style. Content is thought of as "what" is depicted and identified with the subject matter, semantics, or subjects indexically referenced, while "style" is thought of as "how" subjects are rendered through the choice of media and techniques the reflect the process of production or individual perspective. Style is associated with genres, time periods, and individual artists. A renaissance painter from Florence and an early 20th century Spaniard might both paint a portrait, but each brings a unique perspective to the subject matter through their different styles.

Content and style are independent in the sense that any style can be combined with any content, and thus their combination can be considered one type of "synthesis" across domains. Synthesis is a characteristic that has been considered as a component of creativity across many fields including visual psychology (e.g. Finke 1988). Conceptual Blending is one approach to formal computational models addressing synthesis in the Computational Creativity arena (Pereira and Cardoso 2003).

Gatys, Ecker, and Bethge (2016) showed that the 19-layer VGG-Network (Simonyan and Zisserman 2014) trained to recognized objects learns representations that can be used to separate content from style, as well as for the generation of arbitrary combinations of style and content (Figure 2).

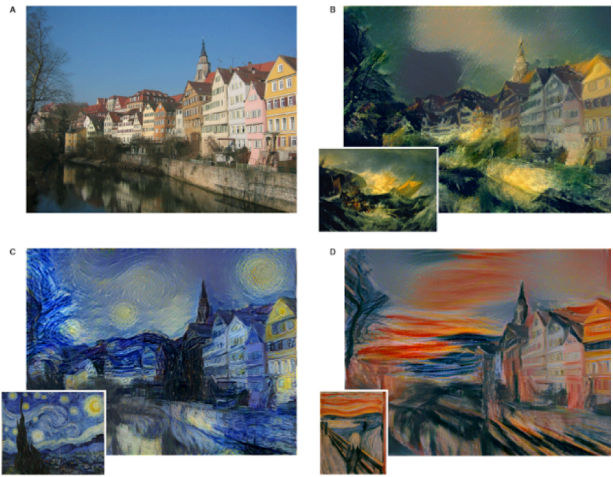

Figure 2. (A) Original photo by Andreas Praefcke rendered in the styles from (B). The Shipwreck of the Minotaur by J.M.W Turner, 1805, (C) The Stary Night by Vincent van Gogh, 1889, and (D) Der Schrei by Edvard Munch, 1893. (Image from Gatys, Ecker, and Bethge (2016) used with permission.)

The mechanism for synthesizing combinations of style from one specific image, and content from another developed by Gatys et al. is reminiscent of the DeapDream network in that during the generative process, an image is presented to the network and then slowly manipulated through the backpropagation of error through the network until it causes activations of the hidden layers to achieve a certain objective. In the case of style transfer however, the objective function for computing error comes in two parts: one is the "content" objective, which is to have the activations at specified layers in response to the formative generated image match the values of those same units in response to the original content image. The second component of the objective is for style. However, rather than trying to achieve a match with actual activation levels of specific hidden layers resulting from the style source image and the generated image, the input is manipulated until there is a match of a statistical measure derived from patterns of activation. The objective based on this second order measure (known as a Gram matrix) maintains correlations between features, but is independent of their spatial location in the 2D representation of the image at the given layers. It just so happens that this measure of statistical correlation between features aligns very closely (although not always completely) with our sense of painterly style as can be seen in the images in Figure 2.

Similar architectures have been shown to work on audio represented as a 2D spectrogram image (Ulyonov and Lebedev 2016), although the results are not as compelling in the audio domain (Shahrin and Wyse 2019). Both Ustyuzhaninov et al. (2016) and Ulyonov and Lebedev (2016) have reported a fascinating aspect of this technique of cross-domain synthesis: that it makes little difference whether or not the network was trained before being used for generation. A shallow network with untrained (randomized) weights can achieve similarly convincing blends of style and content from different images. That is, it turns out the architecture itself, along with the statistical style measure, and the dual style-plus-content objective function, are sufficient for this process of cross-domain synthesis. Note that the link between perceptual (input) process and generation is still present in this architecture because image generation is achieved through the back propagation of error through the same network that responds to image input. Although we might not recognize the process as apophenia without the influence of the learning that we saw in the DeepDream network, the technical process is still one of "reading in" through the mechanism of backpropagation to manipulate an image until it meets criteria defined by an internal representation of input images.

## Sentiment

While sentiment, emotion, and affect are not generally considered as defining creativity, they are certainly associated with motivations, processes, and reception of creative artistic works. Understanding emotion is an important part of human social communication, and thus for computational perceptual systems such as face and voice recognition as well. The automatic or parametric control of systems to induce particular emotional responses has been the focus of some generative games (Freeman 2004), and music (Livingston et al. 2007). Picard (1997) considered imbuing computers with the ability to express, recognize, and exhibit emotional behavior (if not to actually "feel" them).

One way to make DNN architectures capable of both categorization and generation of affect would be simply to start with a massive data set with affective labels for supervised training, and hope that there is statistical consistency in the data that the machine can detect. This section discusses a system that developed a kind of "emotional intelligence" without being explicitly trained to do so.

This story starts with a Recurrent Neural Network trained to generate natural language reviews of products. Recurrent neural networks lend themselves to learning and generating sequential data such as speech, music, and text. Data is typically streamed as input to the network one token at a time (e.g. characters from the text of a product review), possibly along with "conditioning" input (such as data from images of products) and the system learns to predict the next character in a sequence (Figure 3).

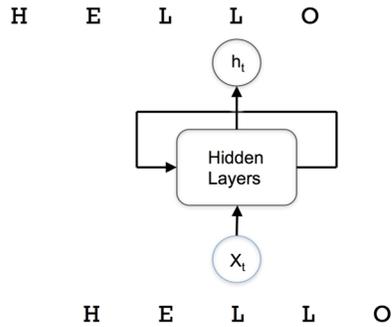

Figure 3. An RNN with recurrent connections training to predict a sequence of characters. The output (top) predicts the next character given the input (bottom), one character per time step.

An RNN can be run in two phases, one as training when the target output is provided by the "teacher" while the network adjusts weights (along each of the arrows in Figure 3), and the other as generation when, after learning, the system produces the next token (in this case a character) in the sequence at the output which is in turn fed back in to the network as the input for prediction at the next step. Karpathy (2015) discusses and interprets creative capabilities of character-generating RNNs.

Radford, Jozefowicz, and Stskever (2017) were interested in studying what kinds of internal language representations an RNN learns in order to accomplish its predictive task. Specifically, they wondered if the learned representation disentangled high-level concepts such as sentiment and semantic relatedness. They approached the problem as a kind of "transfer learning" paradigm where training on one task (character prediction) is used as a starting point for learning a different task (e.g. sentiment identification).

Radford et al. first trained their predictive system on Amazon product reviews. After training, the RNN can be run in generative mode using each predicted next character output as input at the next time step. It thereby learns to spin text resembling product reviews.

They then consider the trained network as a language model. To test whether the language model internally represents recognizable high-level concepts, a separate linear classifier was trained on the state of the penultimate hidden layer activations during and after processing a product review. Taking sentiment as a particular example, a linear classifier was trained to map this input to the externally provided (supervised) "positive" or "negative" assessment. The performance of the linear classifier is interpreted as the measure of how well the concept (in this case, sentiment) was captured by the original predictive RNN in the representation of the review.

They discovered that not only was the network able to achieve state-of-the-art performance on rating reviews as positive or negative (compared with networks that had been fully trained specifically for that task), they furthermore noticed that a single node in the representation was responsible for the decision about sentiment. That is, the RNN that was trained only to generate successive characters in a review learned to represent sentiment as a value of a single node in the network. The activation response of that node shows a very clear bimodal distribution depending on the actual sentiment of the review (Figure 4).

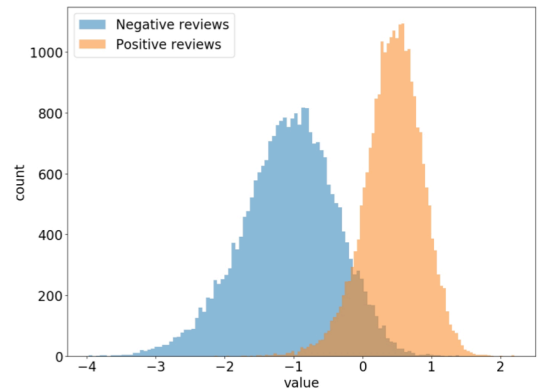

Figure 4. The activation of a single unit in the penultimate layer of a predictive RNN shows a clear binmodal activation pattern corresponding to the sentiment of the review. (Image from Radford, Jozefowicz, and Stskever (2017) used with permission).

The unsupervised character-level training also provides insight into another important dimension of creative computation in deep learning architectures – the ability of such systems to learn for themselves what aspects of an unstructured environment are significant. That is, it discovered the value of the sentiment representation, the meaning of which is grounded only in its own separate predictive learning task.

Furthermore, the sentiment node can be utilized parametrically in the generative phase of the RNN. By "clamping" it to a value representing the negative or positive assessment and letting the RNN run its character-by-character review synthesis, the reviews it constructs can be clearly recognized as having the desired sentiment characteristic despite all the other variation in a product review.

## Analogy and Metaphor

The late 17th century philosopher Giambatista Vico, in his 1725 *The New Science*, recognized the role of metaphor as the foundation of language, reasoning that figurative language precedes literal language and that word meaning only becomes fixed through convention. Some three hundred years later, word2vec models (Mikolov 2013) took a bold step in the field of language modeling to represent words based only on the context (consisting of other words) in which they appear.

Some sort of vector representation of words is typically used to feed input to neural networks. One way to motivate and understand the word2vec strategy is to start from a "one-hot" baseline representation. A one-hot representation is a vector with a length equal to the number of possible different items to be represented. In this case, that length is equal to the number of words in the vocabulary. Each word is represented with a '1' in its unique position in the vector, and '0's elsewhere. For example, for two particular words we might have:

scary: … 0, 0, 0, 0, 0, 0, 0, 0, 0, 0, 0, 0, 0, 0, 1, 0, 0, 0, 0, 0, …
dog:   … 0, 0, 1, 0, 0, 0, 0, 0, 0, 0, 0, 0, 0, 0, 0, 0, 0, 0, 0, 0, …

The advantage of one-hot coding is the direct dictionary-like mapping between the vector and the word. However, two issues are immediately apparent. One is the inefficiency of a vocabulary-length vector for each word. The other issue is that the representation does not capture any of the structure that might be inherent in the data such as semantic relatedness or syntactic similarities. To address at least the first problem, a "distributed" code would be much more memory efficient, decreasing the length of each vector, but having more non-zero values in the vector for each word.

A typical way to reduce the dimension of a representation for neural networks is to train an "autoregressive" neural network to learn a reduced representation. An autoregressive neural network simply learns to reproduce its input at the output, but has a hidden layer that is of much lower dimension than the input and output dimensions. By training the network to reproduce all the one-hot representations, the activations in a hidden layer can be interpreted as a new compact distributed representation for the words. The autoregressive network can then function as the encoder (mapping one-hot to distributed representations) to produce the code for use in other neural network tasks and architectures, and as a decoder (mapping the distributed representation back to the one-hot representation to retrieve the human-readable word.

If it were possible to additionally endow the compact distributed codes with some sense of the meanings of the words, then the representation might further assist the networks that will be doing the language processing. The word2vec "Continuous Bag of Words" (CBOW) technique uses the same idea as the autoencoder for reducing dimension, but instead of learning to map just the single word to itself, it learns to map all n-word sequences in a database that surround the target word to the word itself at the output (Figure 5). For example, "The dog barked scaring the burglar away" and "The child enjoyed scaring neighbors on Halloween." would be two of the (many) thousands of contexts containing "scaring" that would be used to learn to produce the word "scaring" at the output. It is easy to construct the input to train the representations just by adding the one-hot vectors of the context words together. No sense of ordering is preserved (thus the derivation of the name of the CBOW technique from "bag of words.")

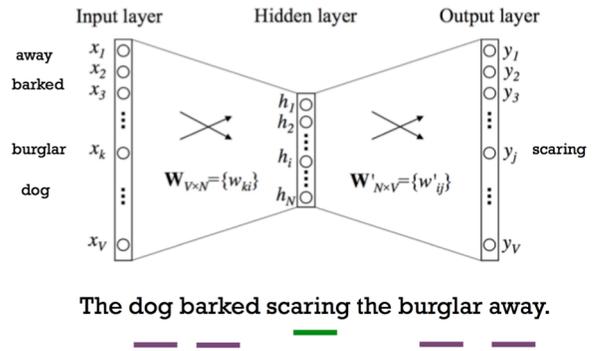

Figure 5. Training a compact distributed representation (the hidden layer) of the word "scaring" using contexts in which the target word appears.

The encoding and decoding function is preserved by this training strategy. However, now the hidden layer code for a given word not only indexes the individual words, but also embeds information about how the word is used in the language.

This representation is beneficial for training neural networks on a wide variety of natural language tasks. However one of the most impressive demonstrations of the elegance of the representation comes from the original Mikolov et al. (2013). As an $m$-dimensional vector, the distributed representation of a word is a point in an $m$-dimensional space where $m$ is much lower than the size of the vocabulary. The question is: does the data have interesting structure in this space, and if so, what kind of structure is it?

It is probably of no surprise that words that have similar semantics occupy nearby points in the representation space based on their similar usage patterns. For example, "frightening" and "scaring" show up in many of the same usage patterns. We can substitute one word for the other in the above example without drastically changing the meaning of the sentences. Because the contexts are similar, the hidden layer learns similar representations.

Simple vector math can be used to explore whether vector operations have any interpretation in terms of the language. For example, taking the vector difference between the points that represent "puppy" and "dog" yields a vector that connects the two points and in some sense "defines" their relationship. What Mikolov demonstrated is that these difference vectors do have semantic meaning. It turns out if we take the same vector that represents the difference between "dog" and "puppy", and place one endpoint on the point representing "cat", the other endpoint lands near the point for "kitten" (Figure 6). This ability of vector relationships to capture semantic relationship provides the means for having the system fill in the blanks for A::B as C::? – a kind of analogical reasoning.

The network wasn't trained for this purpose explicitly, but it self organizes given the task of learning word repre-sentations based on usage context. Ha and Eck (2017) have done the same kind of vector math on latent vector representations for analogical generation in the domain of drawing images. The Google Magenta group has also done similar work in the domain of music. (Roberts et al. 2918).

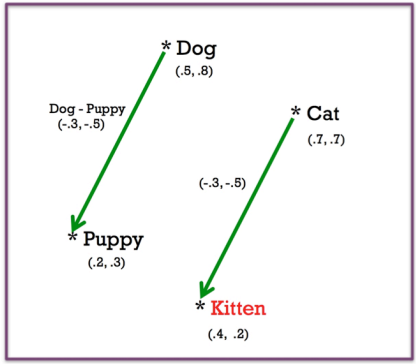

Figure 6. The vector between points representing Dog and Puppy, (-.3, -.5), when taken from the point representing Cat, points to the neighborhood of Kitten.

## Abstraction

Philosopher Cameron Buckner's (2018) study of deep convolutional networks (DCNN) mechanisms produced a notion of "transformation abstraction." The layers, convolutions, and pooling mechanisms of DCNNs "jointly implement a form of hierarchical abstraction that reduces the complexity of a problem's feature space (and avoids overfitting the network's training samples) by iteratively transforming it into a simplified representational format that preserves and accentuates task-relevant features while controlling for nuisance variation." That is, classification in DNNs is identified with a process of abstraction.

In the realm of art, there are at least two types of images we call "abstract." One is comprised of shapes, lines, and colors that are not readily identifiable as any real-world object. Examples include the cube and line "neoplastic" paintings of Mondrian. A second kind of abstract image is comprised of some features that form the basis for an identification as a real-world object, but includes others that are generally not associated with the real world object, and lacks many that are. Duchamp's Nude Descending a Staircase No. 2 can be considered of this type.

Tom White's *Perception Engine* (2018 a, b) foregrounds the notion of abstraction as well as the role of perception in the generative process. One component of the Engine is an image generator constrained to generate combinations of curved lines and blobs. The other is a pre-trained classification network trained on many thousands of example images, each belonging to one of a large collection of labeled categories (e.g. electric fan, golf cart, basketball).

The generator starts with a random combination of the shapes it is constrained to draw, and iteratively manipulates the image to increase the strength of a specific category in a classification network. Figure 7 shows an image generated to maximize the "hammerhead shark" category.

White's system exposes regions of learned categories in the classifiers where no training examples were provided. The spaces of images that the categorizer classifies together are not otherwise apparent from examining the trained network or the images in the dataset it was trained on.

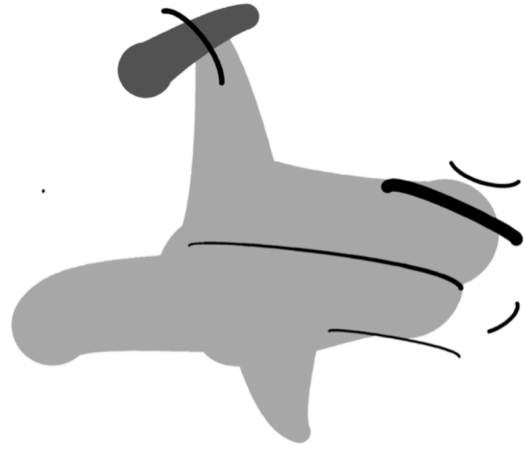

Figure 7. Tom An image from Tom White's Perception Engine that gets classified as a hammerhead (shark).

The shapes of learned categorical regions can be surprising, and often confront researchers with serious challenges. For example, most trained classifiers, even those that appear to be accurate and generalize well on natural images, are subject to being fooled (Nguyen et al. 2015). "Adversarial examples" (Szegedy et al. 2013), are derived from images that would be classified correctly, but that are modified slightly in a way typically imperceptible to a human, sometimes by only a single pixel (Su et al. 2019), causing misclassification of the modified image.

Adversarial image research focuses on images very near to classification boundaries and reveals the formation of categories very different from those of humans. Tom White's system on the other hand, explores the weird and wonderful high-dimensional spaces far from boundaries that the system has learned by generalizing during training.

The Perception Engine starts with an unidentifiable "abstract" image (by virtue of its pallet and the randomness), and finishes with an abstract image that contains some local and/or structural features sufficient to generate strong categorical responses in the classifier, while entirely lacking others that real-world objects (including those used to train the networks) have. White say that "the system is expressing its knowledge of global structure independent of surface features or textures" which aligns with the use of the term abstraction. He has also shown that it is sometimes possible for humans to identify the same classification label as do the neural networks, or at least to be able to recognize features in the image that might have caused the neural network classification after its selection is known.

One of the most remarkable features of this system becomes apparent when the resulting generative works are shown to many different deep-learning classifiers that have been trained on the same real-world data. Despite their different network architectures, the different systems tend to make the same classifications on White's abstractions. That is, if one network classifies the Perception Engine blob-and-line drawing as a "fan", then the image tends to strongly activate the same "fan" category in other classifiers. Generalization patterns in neural networks appear to be general. As a further indication of the nature of the "abstraction" capabilities the networks exhibit, one of his images was generated by optimizing its score for the category of a "tick" insect on 4 neural networks. The image then scores more highly in the tick category than images from the actual human-labeled class of ticks used for validation during training.

It is also interesting to consider the Perception Engine as a kind of "style" machine. The constraints, or 'artistic discipline' of the generator define a style in terms of the types of shapes and lines the system will use to construct images. The target categories can be considered "content," and a series of different drawings by the same generator of different categorical targets would clearly be recognizable as different content sharing a "style," as are most of the works produced by the Perception Engine.

## Summary

Taken together, the mechanisms considered in this paper that support the 5 different example behaviors associated with creativity reveal some patterns. The behaviors exhibited by deep learning neural networks are not explicitly programmed, but rather emerge from the simple interaction programmed at the level of nodes and weights. Furthermore, many of the emergent mechanisms identified that serve creative purposes in generative networks arise during training on entirely different and often non-generative tasks. A classification task produced generalization capabilities exploited for abstract image synthesis, a prediction task led to a sentiment representation that afforded parametric control over affect in text generation, and a representational efficiency lent itself to the ability to discover semantic relationships and construct analogies.

Although a wide variety of network types were considered here, all of them use machinery designed for the perceptual processing of media in some way. Some generated output using the exact same substrate used for perception (the DeapDream network) while others used separate, but intimately collaborating perceptual and generative systems (e.g. White's Perception engine). Perceptual capabilities are at the core of deep learning networks, and this paper has illustrated the richness of the connections to creative generation that come from specific computational mechanisms comprising these systems. Perhaps the richness of the connections should come as no surprise given our understanding of human creativity. As Michelangelo wrote "we create by perceiving and that perception itself is an act of imagination and is the stuff of creativity."

## Acknowledgements

This research was supported by a Singapore MOE Tier 2 grant, "Learning Generative and Parameterized Interactive Sequence Models with Recurrent Neural Networks," (MOE2018-T2-2-127), and by an NVIDIA Corporation Academic Programs GPU grant.